\documentclass{article}

\usepackage{arxiv}

\usepackage[utf8]{inputenc} 
\usepackage[T1]{fontenc}    
\usepackage{hyperref}       
\usepackage{url}            
\usepackage{booktabs}       
\usepackage{amsfonts}       
\usepackage{amsmath}
\usepackage{amssymb}
\usepackage{bm}
\usepackage{nicefrac}       
\usepackage{microtype}      
\usepackage{graphicx}

\title{GNBAN: Graph Neural Basis Attention Networks for Long-Horizon Forecasting over Large Entity Sets}

\author{
 Janak M. Patel\thanks{Corresponding Author: \texttt{janak.patel@quantiphi.com}} \\
  Phi Labs, Quantiphi\\
  Marlborough, MA 01752, USA \\
  \And
 Anirudh Deodhar \\
  Phi Labs, Quantiphi\\
  Marlborough, MA 01752, USA \\
  \And
 Dagnachew Birru \\
  Phi Labs, Quantiphi\\
  Marlborough, MA 01752, USA \\
}

\begin{document}
\maketitle
\begin{abstract}
Demand forecasting at the lowest level of a retail hierarchy requires simultaneously predicting tens of thousands of correlated long-horizon time series across products, stores, and regions. This creates a fundamental challenge for modern forecasting systems: models must scale across massive retail catalogs while capturing shared demand dynamics and producing forecasts that practitioners can interpret and trust. Classical statistical methods require maintaining separate models for each series, making them difficult to manage at production scale. Deep auto-regressive architectures struggle as the joint forecasting state grows to tens of thousands of dimensions. More recently, graph-based deep forecasting architectures have shown promise in modeling dependencies across entities and time. However, many existing approaches remain difficult to scale for long-horizon forecasting and often generate opaque forecasts that provide limited insight into underlying demand patterns. To address these challenges, we propose \textbf{GNBAN (Graph Neural Basis Attention Network)}, an end-to-end forecasting architecture that combines heterogeneous graph representation learning with an interpretable basis-decomposition forecasting head. Retail data are represented directly as a heterogeneous graph derived from the underlying relational database schema, allowing a single model to learn across the entire catalog. Instead of predicting forecast horizons directly, GNBAN decomposes each forecast into three semantically meaningful components: trend, seasonal, and generic. The key innovation is a \emph{per-basis attention mechanism} in which each basis function maintains its own learnable query and retrieves information independently from the historical neighborhood of the forecast entity. This enables different bases to specialize to distinct temporal patterns while preserving interpretability. We evaluate GNBAN on two large-scale retail forecasting benchmarks, M5 Walmart and Favorita Grocery Sales, under matched experimental protocols. Compared with a baseline graph forecasting model, GNBAN improves volume-weighted WRMSSE by approximately 4--5\% across both datasets. Qualitative analysis further shows that the learned decomposition produces interpretable forecasts that expose trend, seasonal, and residual demand drivers without requiring post-hoc explanation methods. These results demonstrate that scalable relational forecasting and interpretable forecast decomposition can be achieved simultaneously within a unified graph-based forecasting framework.
\end{abstract}

\section{Introduction}

Retail demand forecasting at the lowest level of the merchandising hierarchy, where future demand must be predicted for every SKU at every store across multi-week horizons, is one of the most challenging problems in large-scale time-series forecasting. Modern retailers such as Corporaci\'on Favorita and Walmart must simultaneously forecast tens of thousands of correlated product-store series within a single operational pipeline~\cite{makridakis2022m5results,favorita2018}. These forecasting systems must satisfy four competing requirements: they must scale across massive catalogs, capture shared demand dynamics across products and stores, support long-horizon forecasting, and remain interpretable enough for operational decision-making.

Existing forecasting paradigms satisfy only subsets of these requirements. Classical statistical methods such as Error-Trend-Seasonality (ETS) and ARIMA provide interpretable forecasts, but typically require maintaining separate models for every series, making deployment and monitoring difficult at catalog scale~\cite{hyndman2021fpp3}. Deep autoregressive architectures such as DeepAR and Temporal Fusion Transformers replace thousands of local models with a single global network~\cite{salinas2020deepar,lim2021tft}, enabling parameter sharing across series. However, they do not explicitly model relational dependencies between entities such as products, stores, and categories when forecasting large retail panels. Gradient-boosted tree approaches, which dominated the M5 competition~\cite{makridakis2022m5results}, improve scalability and predictive accuracy through extensive feature engineering, yet often flatten temporal and cross-entity relationships into tabular representations, limiting their ability to naturally capture interactions across the retail ecosystem.

More recently, graph-based forecasting methods and relational deep learning approaches~\cite{fey2023rdl,robinson2024relbench} have emerged as a promising direction for large-scale retail forecasting. By representing products, stores, categories, and temporal events as nodes in a heterogeneous graph, these approaches naturally capture cross-entity dependencies through message passing. However, existing graph forecasting architectures typically attach a simple multi-layer perceptron forecasting head to the learned graph embeddings, directly mapping latent representations to forecast vectors. While effective for representation learning, these forecasting heads remain poorly suited for long-horizon retail forecasting: they often produce opaque forecasts, provide little insight into trend or seasonal structure, and degrade as forecasting horizons grow longer.

In parallel, long-horizon forecasting research has shown that decomposition-based architectures can significantly improve forecasting stability and interpretability by explicitly modeling low-frequency structures such as trend and seasonality~\cite{oreshkin2020nbeats,challu2023nhits,wu2021autoformer}. However, these methods are primarily designed for independent time series and do not leverage the rich relational structure present in large retail systems.

This work addresses the gap between scalable graph-based forecasting and accurate, interpretable long-horizon forecasting. We propose \textbf{GNBAN (Graph Neural Basis Attention Network)}, a single global forecasting architecture that combines graph-based representation learning with interpretable basis-style forecast decomposition. GNBAN couples a heterogeneous graph neural network backbone with a basis-attention forecasting head that decomposes each forecast into three semantically meaningful components: a smooth trend component, a periodic seasonal component, and a residual generic component. Each component is generated using a dedicated learnable attention query over the entity's historical neighborhood, enabling the model to capture shared demand dynamics while improving long-horizon forecasting performance and exposing the underlying structure of the forecast. A single trained GNBAN model serves the entire retail catalog, eliminating the need for thousands of independently maintained forecasting models while simultaneously supporting scalable forecasting, interpretable forecast decomposition, and long-horizon prediction within a unified end-to-end framework. Our contributions are summarized as follows:

\begin{enumerate}
\item We introduce GNBAN, a single global forecasting architecture for large-scale retail demand forecasting that combines graph-based representation learning with basis-style forecast decomposition.

\item We propose a basis-attention forecasting head that decomposes forecasts into interpretable trend, seasonal, and residual components while improving long-horizon forecasting performance.

\item We demonstrate that a single trained model can simultaneously achieve scalable forecasting, improved predictive accuracy, and interpretable forecast decomposition across large retail forecasting benchmarks.
\end{enumerate}

\section{Related Work}
\label{sec:related}

\textbf{Large-scale retail demand forecasting.}
Retail forecasting has been extensively studied through the M-series forecasting competitions, particularly M5, which focuses on hierarchical retail demand forecasting across thousands of product-store combinations~\cite{makridakis2022m5results,makridakis2020m4}. Classical statistical methods such as ARIMA, ETS, and Theta remain competitive for individual time series due to their simplicity and interpretability~\cite{hyndman2021fpp3}. However, maintaining separate models for tens of thousands of retail series becomes operationally challenging on a catalog scale. Consequently, recent state-of-the-art solutions increasingly rely on global learning approaches that share information across related series.

\textbf{Global forecasting models.}
Global forecasting architectures learn a single model across all time series in a dataset. DeepAR~\cite{salinas2020deepar}, DeepFactors~\cite{wang2019deepfactors}, and Temporal Fusion Transformers~\cite{lim2021tft} demonstrated that parameter sharing can significantly improve forecasting performance while reducing maintenance overhead. More recently, long-horizon forecasting architectures such as Informer~\cite{zhou2021informer}, Autoformer~\cite{wu2021autoformer}, PatchTST~\cite{nie2023patchtst}, and TimesNet~\cite{wu2022timesnet} have improved the modeling of long-range temporal dependencies. However, these approaches primarily focus on temporal dynamics and do not explicitly model relationships among products, stores, and other entities present in retail systems.

\textbf{Graph-based forecasting.}
Graph neural networks provide a natural mechanism for modeling dependencies across related entities. Existing graph forecasting approaches such as STGCN~\cite{yu2018stgcn}, Graph WaveNet~\cite{wu2019graphwavenet}, MTGNN~\cite{wu2020mtgnn}, and AGCRN~\cite{bai2020agcrn} learn graph structures directly from correlated time series and have shown strong performance on traffic and sensor forecasting tasks. More recently, graph-based learning on relational databases has emerged as a promising framework for enterprise forecasting applications~\cite{fey2023rdl,robinson2024relbench}, where products, stores, categories, and transactions can be represented as nodes connected through natural business relationships. While these approaches effectively capture cross-entity dependencies, forecasting is typically performed using simple MLP-based prediction heads that directly map learned representations to future values.

\textbf{Forecast decomposition and interpretability.}
Forecast decomposition has proven highly effective for long-horizon forecasting. N-BEATS~\cite{oreshkin2020nbeats} introduced interpretable trend and seasonality bases, while N-HiTS~\cite{challu2023nhits} extended this idea through hierarchical interpolation. Similarly, decomposition-based transformer architectures such as Autoformer~\cite{wu2021autoformer} explicitly separate trend and seasonal patterns to improve long-range forecasting performance. Despite their success, these methods are primarily designed for independent time series and do not leverage the relational structure commonly found in large retail systems.

\textbf{Positioning of GNBAN.}
GNBAN bridges these research directions by combining graph-based forecasting with basis-style forecast decomposition. Unlike existing graph forecasting models that rely on black-box prediction heads, GNBAN introduces a basis-attention forecasting head that decomposes forecasts into interpretable trend, seasonal, and residual components while leveraging graph-based representations to capture cross-entity demand dynamics.
\section{Method}
\label{sec:method}

\begin{figure*}[t]
  \centering
  \includegraphics[width=0.95\linewidth]{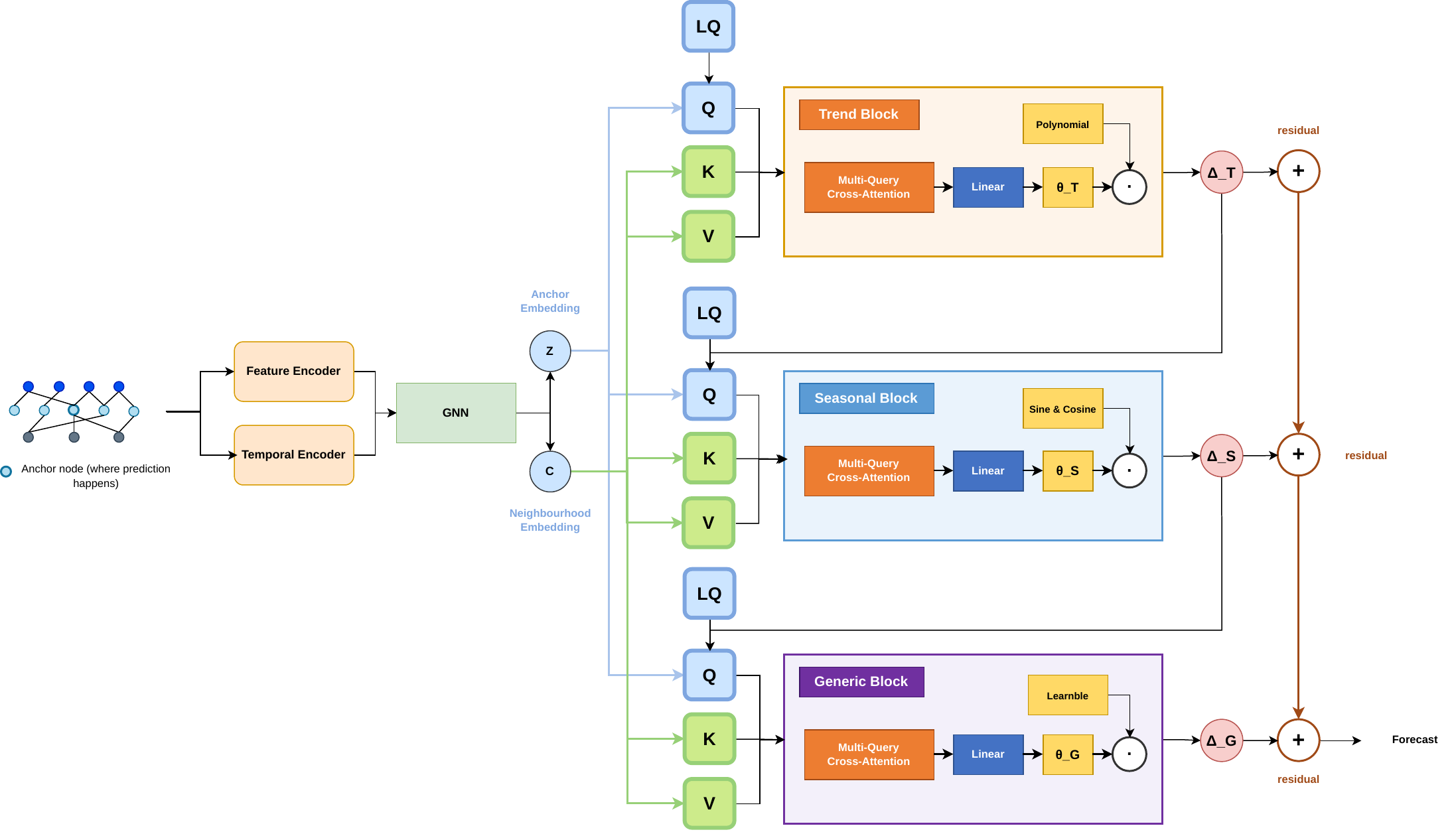}
  \caption{Overview of GNBAN. The retail database is represented as a heterogeneous graph where each forecast entity is connected to its historical sales observations. A heterogeneous GraphSAGE produces an entity embedding and contextual neighborhood representations. The basis-attention forecasting head then generates three additive forecast components: trend, seasonal, and generic using dedicated learnable queries that attend to the historical neighborhood. These components are combined to produce the final long-horizon forecast.}
  \label{fig:architecture}
\end{figure*}

GNBAN (\textbf{G}raph \textbf{N}eural \textbf{B}asis \textbf{A}ttention \textbf{N}etwork) is a single global forecasting architecture designed for large-scale retail demand forecasting. The model combines a heterogeneous graph encoder that captures dependencies among retail entities with a basis-attention forecasting head that produces interpretable long-horizon forecasts. Figure~\ref{fig:architecture} provides an overview of the architecture.

The workflow consists of four stages. First, the retail database is represented as a heterogeneous graph derived from the underlying relational schema. Second, a heterogeneous graph neural network encodes each forecast entity and its historical context into latent representations. Third, a basis-attention forecasting head generates the forecast as an additive combination of trend, seasonal, and residual components. Finally, the entire architecture is trained end-to-end using a normalized forecasting objective.

\subsection{Graph Representation}
\label{ssec:graph}

We represent the retail dataset as a heterogeneous graph constructed directly from the underlying relational schema. The graph contains two node types:

\begin{itemize}
\item \texttt{itemstore}: one node for each SKU-store forecasting entity.
\item \texttt{sales}: one node for each historical observation associated with an entity.
\end{itemize}

Item-store nodes contain static attributes such as product category, department, store identifier, and geographic information. Sales nodes contain historical demand values and time-varying covariates including price, promotions, holidays, and calendar features.

Edges follow the natural foreign-key relationship between sales observations and their corresponding item-store entity. For each forecast entity, we sample its historical sales neighborhood prior to the forecast origin, ensuring that no future information is available during training or inference.

\subsection{Heterogeneous Graph Encoder}
\label{ssec:gnn}

Given a target item-store entity $v$ and its sampled neighborhood $\mathcal{N}(v)$, a heterogeneous GraphSAGE encoder aggregates information from neighboring sales observations to produce contextual representations. Let $\mathbf{x}_v$ denote the feature vector of entity $v$ and $\mathbf{x}_u$ denote the feature vector of neighboring sales node $u$. The encoder computes

\begin{equation}
\mathbf{z}_v
=
\sigma
\left(
W_{\mathrm{self}}\mathbf{x}_v
+
\mathrm{Mean}_{u\in\mathcal{N}(v)}
W_{\mathrm{nbr}}\mathbf{x}_u
\right)
\end{equation}

where $\sigma(\cdot)$ denotes a non-linear transformation followed by layer normalization. The resulting embedding $\mathbf{z}_v \in \mathbb{R}^{d}$ captures both entity-specific information and historical demand context. In addition to the target embedding, the encoder produces contextual embeddings

\[
\mathbf C_v =
[\mathbf h_1,\ldots,\mathbf h_N]
\in \mathbb R^{N \times d}
\]

where $N$ is the number of neighboring sales nodes and $d$ is the hidden dimension. 
\subsection{Basis-Attention Forecasting Head}
\label{ssec:head}

The basis-attention forecasting head is the key contribution of GNBAN. Rather than directly mapping latent representations to an $H$-step forecast vector through a multilayer perceptron, GNBAN represents forecasts as a weighted combination of interpretable basis functions.

\subsubsection{Forecast Decomposition}

The forecasting horizon is represented using three groups of basis functions:

\begin{itemize}
\item \textbf{Trend basis}: polynomial functions capturing long-term demand evolution.
\item \textbf{Seasonal basis}: sinusoidal functions capturing recurring demand cycles.
\item \textbf{Generic basis}: learnable basis functions capturing residual structure.
\end{itemize}

Let

\begin{equation}
\mathbf{B}
=
\left[
\mathbf{B}^{(T)},
\mathbf{B}^{(S)},
\mathbf{B}^{(G)}
\right]
\end{equation}

denote the complete basis dictionary defined over the forecasting horizon. The final forecast is decomposed as

\begin{equation}
\hat{\mathbf{y}}_v
=
\hat{\mathbf{y}}_v^{(T)}
+
\hat{\mathbf{y}}_v^{(S)}
+
\hat{\mathbf{y}}_v^{(G)}
\label{eq:forecast_decomposition}
\end{equation}

where the three terms correspond to trend, seasonal, and generic contributions, respectively.

\subsubsection{Per-Basis Attention}

Unlike conventional attention mechanisms that use a shared query representation, GNBAN assigns a dedicated learnable query to each basis function. This allows different basis functions to retrieve different aspects of the historical signal. Let $\mathbf{q}_k \in \mathbb{R}^{d}$ denote the learnable query associated with basis $k$. To adapt the behavior of each basis to a specific forecast entity, the queries are conditioned on the entity embedding $\mathbf{z}_v$ produced by the graph encoder:

\begin{equation}
\mathbf{C}^{(q)}_v
=
\mathrm{reshape}
\left(
\mathbf{W}_{\mathrm{cond}}
\mathbf{z}_v
\right)
\in
\mathbb{R}^{K \times d}
\end{equation}

where row $k$ corresponds to the conditioning vector $\mathbf{c}_{v,k}$. The basis-specific query is then

\begin{equation}
\tilde{\mathbf q}_{v,k}
=
\mathbf q_k
+
\mathbf c_{v,k}
\end{equation}

Keys and values are generated from the neighborhood embeddings produced by the graph encoder:

\begin{equation}
\mathbf K_v
=
\mathbf C_v \mathbf W_K,
\qquad
\mathbf V_v
=
\mathbf C_v \mathbf W_V
\end{equation}

The attention weights for basis $k$ are computed as

\begin{equation}
\boldsymbol{\alpha}_{v,k}
=
\mathrm{softmax}
\left(
\frac{
\tilde{\mathbf q}_{v,k}^{\top}
\mathbf K_v^{\top}
}
{\sqrt d}
\right)
\end{equation}

where $\boldsymbol{\alpha}_{v,k}\in\mathbb{R}^{N}$ assigns importance to each historical observation in the neighborhood. The retrieved context vector is

\begin{equation}
\mathbf r_{v,k}
=
\boldsymbol{\alpha}_{v,k}
\mathbf V_v
\end{equation}

A scalar coefficient for basis $k$ is then obtained as

\begin{equation}
\beta_{v,k}
=
\mathbf w_k^\top
\mathrm{LayerNorm}
(\mathbf r_{v,k})
+
b_k
\end{equation}

Because each basis maintains its own query, different basis functions can attend to different regions of the historical neighborhood. Trend bases can focus on long-term demand evolution, seasonal bases can attend to phase-aligned historical observations, and generic bases can specialize to short-term residual behavior.

\subsubsection{Forecast Reconstruction}

The coefficients generated by the attention mechanism weight their corresponding basis functions to produce horizon-level forecasts:

\begin{align}
\hat{\mathbf y}_v^{(T)}
&=
\sum_{k \in \mathcal T}
\beta_{v,k}\mathbf b_k
\\
\hat{\mathbf y}_v^{(S)}
&=
\sum_{k \in \mathcal S}
\beta_{v,k}\mathbf b_k
\\
\hat{\mathbf y}_v^{(G)}
&=
\sum_{k \in \mathcal G}
\beta_{v,k}\mathbf b_k
\end{align}

The final forecast is obtained by applying a softplus activation to ensure non-negative predictions:
\begin{equation}
\hat{\mathbf{y}}_v = \mathrm{softplus}\!\left(\hat{\mathbf{y}}_v^{(T)} + \hat{\mathbf{y}}_v^{(S)} + \hat{\mathbf{y}}_v^{(G)}\right).
\end{equation}
This decomposition provides a direct interpretation of how trend, seasonality, and residual demand dynamics contribute to the prediction.

\section{Experiments}
\label{sec:experiments}

\subsection{Datasets}
\label{sec:datasets}

We evaluate GNBAN on two widely used retail demand forecasting benchmarks that differ substantially in scale, demand characteristics, and forecasting horizon. Together, they provide complementary evaluation scenarios spanning intermittent count-based demand and continuous, heavy-tailed sales distributions.

\textbf{M5 Walmart Accuracy}~\cite{m52020kaggle,makridakis2022m5results} contains 30,490 item-store time series collected from 10 Walmart stores across three U.S.\ states, spanning 1,941 days from January 2011 to June 2016. The dataset exhibits highly intermittent demand, with a large fraction of zero-sales observations and daily sales ranging from 0 to approximately 700 units. In addition to historical sales, M5 provides a 12-level product hierarchy, SNAP eligibility indicators, event calendars, and weekly selling prices. We use data from 2014-01-01 onward ($\approx$26.6M rows after filtering) and evaluate 28-day forecasts at the item-store level using the full set of 30,490 series following the official public leaderboard protocol.

\textbf{Favorita Grocery Sales}~\cite{favorita2018} contains sales records from 54 Corporaci\'on Favorita stores in Ecuador spanning approximately 1,680 days from January 2013 to August 2017. Unlike M5, sales are continuous and span a substantially wider dynamic range, reaching approximately 44,000 units per day. The dataset includes rich exogenous information such as promotions, holidays, crude oil prices, store transaction counts, and product metadata. We use data from 2016-01-01 onward ($\approx$25.9M rows) and retain the 50,000 highest-volume item-store pairs after applying a minimum cumulative sales threshold. We evaluate a 16-day forecasting horizon.

The two datasets differ not only in scale and horizon length, but also in their underlying demand distributions and exogenous drivers, providing a rigorous test of model robustness across diverse retail forecasting environments.

\subsection{Evaluation Protocol and Metrics}
\label{sec:protocol}

\subsubsection{Forecast Origins and Data Splits}

\textbf{M5.} We follow the official M5 public leaderboard protocol, using 2016-04-24 as the forecast origin and generating forecasts for the subsequent 28 days. Evaluation is performed at the bottom level of the hierarchy (item-store pairs) using the complete set of 30,490 series. Training forecast origins are sampled weekly prior to the validation and test periods, while preserving temporal ordering.

\textbf{Favorita.} Since the original competition provides labels for only a single forecasting window, we construct a reproducible evaluation protocol using the final 32 days of available data. Weekly forecast origins are used for training, while non-overlapping validation and test windows are created from the final two 16-day periods. This setup preserves the original forecasting horizon while enabling reliable model selection and evaluation.

All baselines are trained and evaluated using identical data splits, preprocessing procedures, and forecasting horizons. Consequently, performance differences reflect architectural choices rather than variations in experimental setup.

\subsubsection{Preprocessing}

Retail demand spans several orders of magnitude across products and stores. To stabilize optimization, all sales-related numerical features are transformed using a log1p normalization prior to graph construction. Model predictions are converted back to the original sales scale using the inverse transformation before evaluation. Following common practice in large-scale retail forecasting, training samples whose target windows contain more than 50\% zero observations are removed to reduce the influence of extremely sparse demand patterns.

\subsubsection{Evaluation Metrics}
We report the Weighted Root Mean Squared Scaled Error (WRMSSE), the
primary M5 competition metric.
For series $i$ with historical observations $y_{i,1},\ldots,y_{i,T}$ and
$H$-step forecasts $\hat{y}_{i,1},\ldots,\hat{y}_{i,H}$, the per-series
scaled error is

\begin{equation}
\mathrm{RMSSE}_i =
\sqrt{
\frac{\dfrac{1}{H}\displaystyle\sum_{t=1}^{H}
      \left(y_{i,t} - \hat{y}_{i,t}\right)^{2}}
     {\dfrac{1}{T-1}\displaystyle\sum_{t=2}^{T}
      \left(y_{i,t} - y_{i,t-1}\right)^{2}}
}
\label{eq:rmsse}
\end{equation}

where the denominator is the mean squared one-step na\"ive forecast error
computed on the pretest period, serving as a scale normaliser across series
with different sales magnitudes.
$\mathrm{RMSSE}_i < 1$ indicates the model outperforms the na\"ive
last-observed-value baseline. Our \textbf{primary metric} is the volume-weighted aggregate

\begin{equation}
\mathrm{WRMSSE}_{\mathrm{vol}} = \sum_{i=1}^{N} w_i\,\mathrm{RMSSE}_i,
\qquad \sum_{i=1}^{N} w_i = 1,
\label{eq:wrmsse_vol}
\end{equation}

where $w_i$ is proportional to series $i$'s total pretest sales volume,
giving higher weight to economically significant high-volume items.
We additionally report an \textbf{unweighted mean}

\begin{equation}
\mathrm{WRMSSE}_{\mathrm{uw}} = \frac{1}{N}\sum_{i=1}^{N}\mathrm{RMSSE}_i
\label{eq:wrmsse_uw}
\end{equation}

which gives equal weight to every series and is therefore sensitive to
performance on low-volume and intermittent-demand items.
Both metrics are computed in original sales units after reversing any
training-time normalisation; lower values indicate better performance,
and 1.0 corresponds to the na\"ive last-day baseline by construction.

\subsection{Baselines}
\label{sec:baselines}
Our primary baseline is a heterogeneous graph neural network forecaster,
referred to throughout the paper as \textbf{Base Graph}. Base Graph uses
the same graph construction, node features, neighborhood sampling strategy,
GraphSAGE backbone, and training procedure as GNBAN. The only difference is
the forecasting head: the proposed basis-attention module is replaced by a
multi-layer perceptron (MLP) that directly predicts the $H$-step forecast
vector. This controlled comparison isolates the contribution of the
proposed basis-attention forecasting head while keeping all other
components identical.

\subsection{Implementation Details}

All numerical sales features and forecasting targets are transformed using
$\log(1+x)$ prior to training and inverted using $\exp(x)-1$ during
evaluation. This normalization improves numerical stability and allows a
single training configuration to generalize across datasets with widely
different sales scales. Following common practice in retail forecasting,
training windows containing more than 50\% zero-valued targets are excluded
to reduce the impact of highly sparse demand patterns.

The graph encoder consists of a single-layer heterogeneous GraphSAGE
backbone with mean aggregation, hidden dimension $d=64$, and LayerNorm
applied after message passing. For each forecast entity, temporal
neighborhood sampling retrieves up to $K=256$ historical sales nodes from
the one-hop neighborhood. The forecasting head contains 22 basis functions:
4 polynomial trend bases, 10 sinusoidal seasonal bases corresponding to
retail-relevant periods, and 8 learnable generic bases. Each basis is
associated with an independent learnable attention query.

Models are trained end-to-end using the Huber loss with $\delta=1.0$ on
log-transformed targets. Optimization is performed using Adam with gradient
clipping at $\|g\|_2 \leq 1.0$ and early stopping based on validation loss
with a patience of 10 epochs. The batch size is fixed at 128 for both
datasets, while the learning rate is set to $10^{-3}$ for M5 and
$10^{-4}$ for Favorita.

All experiments are conducted on a single NVIDIA V100 GPU (16\,GB). The
heterogeneous graph is stored in CPU memory and sampled on-the-fly, with
only the sampled mini-batch subgraphs transferred to the GPU during
training.

\subsection{Forecasting Performance}
Table~\ref{tab:main} reports the forecasting performance of GNBAN and the matched \textit{Base Graph} baseline on the M5 and Favorita benchmarks. The Base Graph model uses the same relational graph construction, node features, neighborhood sampling strategy, heterogeneous GraphSAGE backbone, and training procedure as GNBAN; the only difference is that the basis-attention forecasting head is replaced with a standard multi-layer perceptron that directly predicts the forecast horizon. GNBAN improves volume-weighted WRMSSE consistently across both datasets. On Favorita, GNBAN reduces $\mathrm{WRMSSE}_{\mathrm{vol}}$ from 0.6741 to 0.6474 ($-4.0\%$) and $\mathrm{WRMSSE}_{\mathrm{uw}}$ from 0.7026 to 0.6695 ($-4.7\%$). On M5, GNBAN improves $\mathrm{WRMSSE}_{\mathrm{vol}}$ from 0.8689 to 0.8372 ($-3.6\%$); the unweighted metric shows no improvement (0.9053 vs.\ 0.9152), as the basis-attention head redistributes model capacity toward high-volume items that dominate the volume-weighted objective. These improvements are particularly noteworthy because they arise solely from the forecasting head. Both models share the same graph representation, message-passing architecture, optimization procedure, and input features. Consequently, the performance gains can be attributed directly to the proposed basis-attention decomposition mechanism rather than increased model capacity or additional information. The results suggest that decomposing forecasts into trend, seasonal, and generic components enables more effective utilization of the historical information encoded by the graph backbone. Rather than predicting the entire forecast horizon through a single latent representation, GNBAN allows different basis functions to retrieve specialized information from the historical neighborhood through independent attention queries, leading to improved long-horizon forecasting performance.

\begin{table}
\centering
\caption{
Bottom-level forecasting performance on the M5 and Favorita benchmarks.
WRMSSE$_{\mathrm{vol}}$ denotes the volume-weighted Weighted Root Mean Squared Scaled Error, while WRMSSE$_{\mathrm{uw}}$ denotes the unweighted mean RMSSE across all item-store series. Both models share identical graph construction, neighborhood sampling, node features, GNN backbone, and training procedure; only the forecasting head differs. Lower values indicate better performance; best results per dataset in bold. $^\dagger$GNBAN vol-WRMSSE improves on M5; UW-WRMSSE does not.
}
\label{tab:main}
\begin{tabular}{llcc}
\toprule
Dataset & Model & WRMSSE$_\text{vol}$ & WRMSSE$_\text{uw}$ \\
\midrule
Favorita         & Base graph            & 0.6741 & 0.7026 \\
(16-day, 50K)    & \textbf{GNBAN} (ours) & \textbf{0.6474} & \textbf{0.6695} \\
\midrule
M5                   & Base graph            & 0.8689 & \textbf{0.9053} \\
(28-day, 30,490)     & \textbf{GNBAN} (ours) & \textbf{0.8372} & 0.9152$^\dagger$ \\
\bottomrule
\end{tabular}
\end{table}

\subsection{Interpretability Analysis}
\label{sec:interpretability}

\begin{figure}[t]
  \centering
  \includegraphics[width=\linewidth]{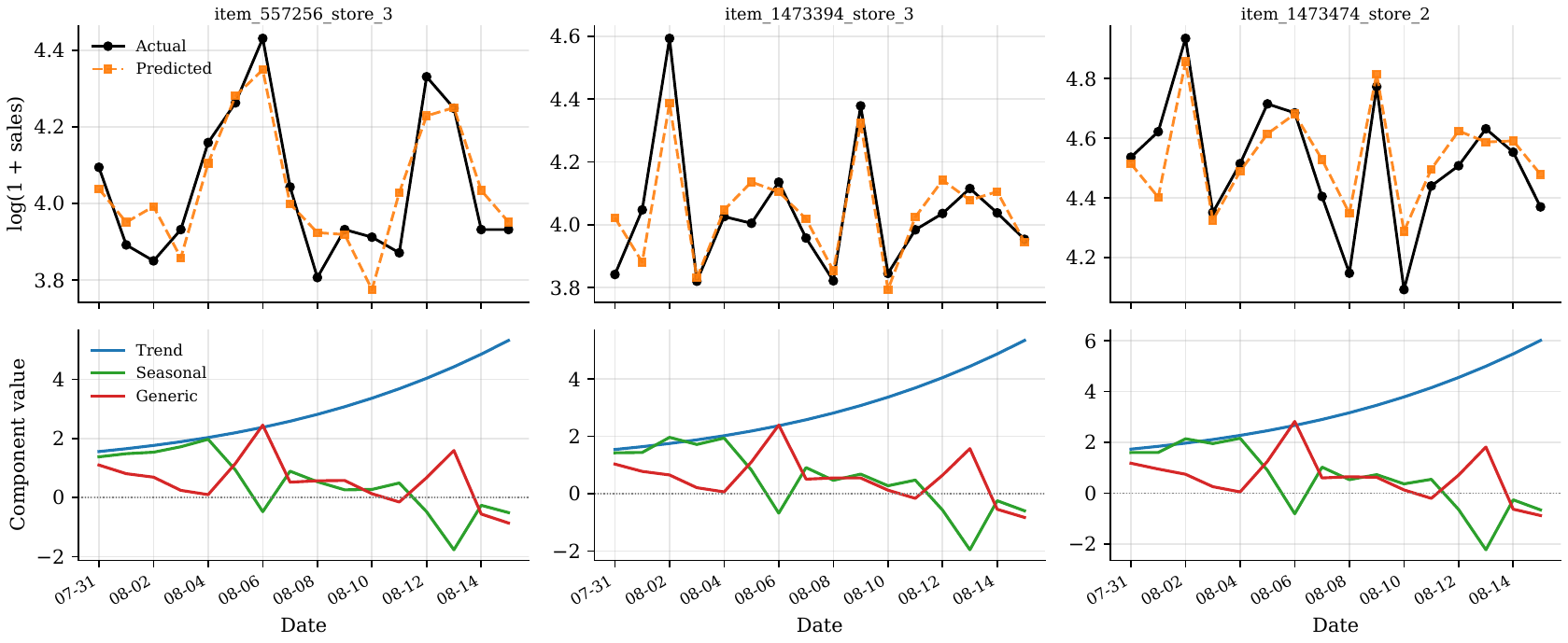}
  \caption{Forecast decomposition for three representative Favorita item-store pairs.
    \textbf{Top:} Actual demand (black) and GNBAN forecast (orange) in log1p space over the 16-day forecasting horizon. The close agreement between the two curves illustrates the forecasting accuracy of the model on these examples.
    \textbf{Bottom:} Decomposition of the forecast into trend, seasonal, and generic components generated by the basis-attention head. The trend component captures long-term demand evolution, the seasonal component captures recurring weekly patterns, and the generic component models residual effects not explained by trend or seasonality. The decomposition is produced directly by the model and requires no post-hoc explanation method.
    }
  \label{fig:decomposition}
\end{figure}

A key advantage of GNBAN is that interpretability is built directly into the forecasting architecture. Rather than producing a forecast through a black-box latent representation, the model explicitly decomposes each prediction into trend, seasonal, and generic components. These components are generated during the forward pass and therefore provide an intrinsic explanation of the forecast without requiring post-hoc interpretation methods such as SHAP or LIME. Figure~\ref{fig:decomposition} presents forecasting results for three representative item-store pairs from the Favorita dataset. The top row compares the actual demand trajectory with the forecast generated by GNBAN over the 16-day prediction horizon. The model accurately captures both the overall demand level and the temporal evolution of each series despite substantial differences in sales magnitude and variability across items. The bottom row shows the corresponding forecast decomposition obtained directly from the basis-attention head:

\begin{equation}
\mathbf{T}_v=\boldsymbol{\theta}^{(t)}_v\mathbf{B}^{(t)},
\qquad
\mathbf{S}_v=\boldsymbol{\theta}^{(s)}_v\mathbf{B}^{(s)},
\qquad
\mathbf{G}_v=\boldsymbol{\theta}^{(g)}_v\mathbf{B}^{(g)}.
\end{equation}

where $\mathbf{T}_v$, $\mathbf{S}_v$, and $\mathbf{G}_v$ denote the trend, seasonal, and generic forecast components, respectively. Several observations emerge. First, the trend component captures the long-term demand trajectory and explains the majority of the forecast magnitude. Second, the seasonal component clearly isolates recurring weekly demand patterns, with both amplitude and phase varying across items. Finally, the generic component captures localized residual effects that cannot be explained by smooth trend or periodic seasonality, such as promotions, short-lived demand spikes, or inventory-related fluctuations.

\begin{figure}[t]
  \centering
  \includegraphics[width=\linewidth]{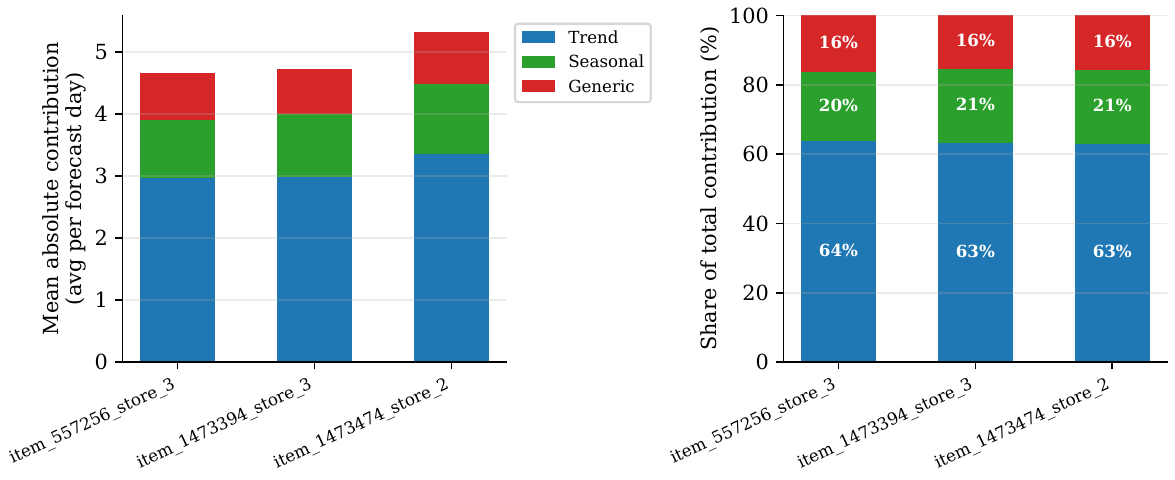}
  \caption{Per-item component contribution magnitudes. \textbf{Left:} Mean absolute contribution per forecast day (log1p units). \textbf{Right:} Percentage share of total contribution. The trend block contributes 63--64\% across all three items; seasonal 20--21\%; generic 16\%. The near-identical proportions across distinct items indicate the model learned a stable, transferable demand structure rather than item-specific noise.}
  \label{fig:contributions}
\end{figure}

To quantify the relative importance of these components, Figure~\ref{fig:contributions} reports their average contribution magnitudes. Across all three examples, the trend component contributes approximately 63--64\% of the total forecast magnitude, the seasonal component contributes 20--21\%, and the generic component contributes roughly 16\%. The consistency of these proportions across different products suggests that the model learns a stable and transferable decomposition of retail demand rather than relying on item-specific memorization.

From an operational perspective, the decomposition provides actionable insights in addition to accurate forecasts. Trend components can support inventory planning and replenishment decisions, seasonal components reveal recurring demand cycles relevant for staffing and logistics, and generic components highlight unusual demand behavior that may warrant further investigation. Consequently, GNBAN offers both improved forecasting performance and a transparent representation of the underlying factors driving each prediction.

\subsection{Ablation Studies}
\label{sec:ablations}
To isolate the contribution of the per-basis attention mechanism, we compare GNBAN against a simplified variant that uses a single attention query per forecast block (trend, seasonal, generic) rather than one per basis function. All other components remain identical. Table~\ref{tab:ablation} summarizes the results.

\begin{table}[h]
\centering
\caption{Ablation on M5: effect of query granularity.}
\label{tab:ablation}
\begin{tabular}{lcc}
\toprule
Variant & WRMSSE$_\text{vol}$ & WRMSSE$_\text{uw}$ \\
\midrule
Base graph (MLP head)              & 0.8689 & \textbf{0.9053} \\
Single query per block (3Q)$^*$    & 0.9173 & 1.2411 \\
\textbf{GNBAN} (per-basis, 22Q)    & \textbf{0.8372} & 0.9152 \\
\bottomrule
\end{tabular}
\end{table}

The single-query variant performs substantially worse than the full GNBAN architecture and is even worse than the MLP baseline on both metrics, with $\mathrm{WRMSSE}_{\mathrm{uw}} = 1.2411 > 1.0$ indicating failure on intermittent-demand items. These results demonstrate that the performance gains are not due to the use of attention alone, but specifically to the ability of individual basis functions to retrieve specialized historical context through independent queries. When all bases within a block share the same query, the retrieved context becomes a single averaged representation, reducing the model's capacity to disentangle long-term trends, recurring seasonal patterns, and residual demand fluctuations independently.

%

\section{Conclusion}
This paper introduces \textbf{GNBAN (Graph Neural Basis Attention Network)}, a forecasting architecture for large-scale retail demand forecasting that combines heterogeneous graph representation learning with interpretable basis-based forecast decomposition. GNBAN represents retail data as a heterogeneous graph and generates forecasts through a basis-attention head that explicitly decomposes predictions into trend, seasonal, and generic components. Experiments on the M5 Walmart and Favorita retail forecasting benchmarks demonstrate that GNBAN consistently improves forecasting accuracy over a matched graph-based baseline while providing intrinsic interpretability. The proposed per-basis attention mechanism enables different forecast components to retrieve specialized information from historical observations, leading to improved long-horizon forecasting performance. Furthermore, the resulting decomposition provides a transparent view of the factors contributing to each prediction, eliminating the need for post-hoc explanation methods. Overall, our results show that accurate forecasting and interpretability need not be competing objectives. By combining graph-based representation learning with basis decomposition, GNBAN provides a unified framework capable of modeling complex retail demand dynamics while producing forecasts that remain understandable to practitioners. Future work will explore larger-scale evaluations against strong forecasting baselines, extensions to additional long-horizon forecasting domains, and transfer-learning settings where models are pre-trained on one retail catalog and adapted to another.


\bibliographystyle{unsrt}
\bibliography{references}

\appendix

\section{Full Hyperparameters}
\label{sec:appendix-hyperparameters}

Table~\ref{tab:full-hyperparameters} lists all architecture, training,
sampling, and preprocessing hyperparameters used in the reported
experiments. Unless otherwise noted, the same configuration is used
for both M5 and Favorita.

\begin{table*}[t]
\caption{Full hyperparameter specification for all reported runs.}
\label{tab:full-hyperparameters}
\centering
\small
\begin{tabular}{p{0.35\linewidth} p{0.58\linewidth}}
\toprule
\textbf{Component} & \textbf{Value} \\
\midrule

\multicolumn{2}{l}{\textit{Graph construction}} \\
Node types &
\texttt{itemstore} (forecast entity), \texttt{sales} \\

Edge type &
PK/FK relation:
\texttt{sales} $\rightarrow$ \texttt{itemstore} \\

Reverse edges &
Added automatically \\

Cross-store same-SKU edges &
Not included (reduced WRMSSE in ablations) \\

Cluster nodes &
Not included (reduced WRMSSE in ablations) \\

\midrule
\multicolumn{2}{l}{\textit{Feature encoding}} \\

Categorical \texttt{itemstore} attributes &
Per-column embedding tables \\

Numerical \texttt{sales} attributes &
Linear encoder with Z-score normalization \\

Multi-categorical event features &
MultiCategoricalEmbedding \\

Temporal encoding &
PyG \texttt{PositionalEncoding} on relative day index \\

\midrule
\multicolumn{2}{l}{\textit{Neighborhood sampling}} \\

Sampler &
PyG \texttt{NeighborLoader} \\

Number of hops &
1 \\

Fanout &
$K=256$ \\

Temporal strategy &
Most-recent-first (\texttt{last}) \\

Future leakage &
Blocked via \texttt{time < seed\_time} constraint \\

\midrule
\multicolumn{2}{l}{\textit{GNN backbone}} \\

Operator &
Heterogeneous GraphSAGE \\

Layers &
1 \\

Aggregation &
Mean \\

Normalization &
LayerNorm \\

Activation &
ReLU \\

Hidden dimension &
$d=64$ \\

\midrule
\multicolumn{2}{l}{\textit{Basis-attention forecasting head}} \\

Trend bases &
4 polynomial bases $(t/H)^k,\; k \in \{0,1,2,3\}$ \\

Seasonal bases &
10 sine/cosine bases with periods
$\{7, 7/2, 7/3, 14, 28\}$ days \\

Generic bases &
8 learnable basis vectors \\

Total basis queries &
22 (one query per basis function) \\

Item conditioning &
Per-basis linear shift derived from seed embedding $\mathbf{z}_v$ \\

Attention scaling &
$1/\sqrt{d}$ \\

Empty-neighborhood handling &
\texttt{nan\_to\_num(0)} \\

Output activation &
Softplus \\

\midrule
\multicolumn{2}{l}{\textit{Target normalization and loss}} \\

Target transform &
$\log(1+x)$ \\

Inverse transform &
$\exp(x)-1$ \\

WRMSSE denominator &
Raw pretest sales series in original scale \\

Loss &
Huber loss ($\delta = 1.0$) \\

NaN protection &
Non-finite batches skipped \\

\midrule
\multicolumn{2}{l}{\textit{Training-table generation}} \\

Forecast origins &
Weekly Mondays \\

Forecast horizon &
M5: 28 days;\quad Favorita: 16 days \\

Sparse-window filter &
Remove windows with $>50\%$ zero targets \\

\midrule
\multicolumn{2}{l}{\textit{Optimization}} \\

Optimizer &
Adam ($\beta_1=0.9,\; \beta_2=0.999$) \\

Learning rate &
$10^{-3}$ (M5), $10^{-4}$ (Favorita) \\

Gradient clipping &
$\|g\|_2 \leq 1.0$ \\

Batch size &
128 \\

Maximum epochs &
50 \\

Early stopping &
Patience = 10 (validation loss) \\

Random seed &
42 \\

\midrule
\multicolumn{2}{l}{\textit{Hardware}} \\

GPU &
1$\times$ NVIDIA V100-SXM2 (16\,GB) \\

Graph storage &
CPU; sampled mini-batches transferred to GPU \\

\bottomrule
\end{tabular}
\end{table*}

\end{document}